# Variational Autoencoder Kernel Interpretation and Selection for Classification

Fábio Mendonça[1,2*], Sheikh Shanawaz Mostafa[2], Fernando Morgado-Dias[1,2], and Antonio G. Ravelo-García[2,3]

[1] University of Madeira, Funchal, Portugal

[2] Interactive Technologies Institute (ITI/ARDITI/LARSyS), Funchal, Portugal

[3] Institute for Technological Development and Innovation in Communications, Universidad de Las Palmas de Gran Canaria, Las Palmas de Gran Canaria, Spain

*Corresponding author; e-mail address: fabio.ruben@staff.uma.pt; postal address: Polo Científico e Tecnológico da Madeira, Caminho da Penteada, piso -2, 9020-105 Funchal Portugal; Phone number: +351 291 721 006*

**Abstract**

This work proposed kernel selection approaches for probabilistic classifiers based on features produced by the convolutional encoder of a variational autoencoder. Particularly, the developed methodologies allow the selection of the most relevant subset of latent variables. In the proposed implementation, each latent variable was sampled from the distribution associated with a single kernel of the last encoder's convolution layer, as an individual distribution was created for each kernel. Therefore, choosing relevant features on the sampled latent variables makes it possible to perform kernel selection, filtering the uninformative features and kernels. Such leads to a reduction in the number of the model's parameters. Both wrapper and filter methods were evaluated for feature selection. The second was of particular relevance as it is based only on the distributions of the kernels. It was assessed by measuring the Kullback-Leibler divergence between all distributions, hypothesizing that the kernels whose distributions are more similar can be discarded. This hypothesis was confirmed since it was observed that the most similar kernels do not convey relevant information and can be removed. As a result, the proposed methodology is suitable for developing applications for resource-constrained devices.

# 1. Introduction

Generative modeling, as produced by *Variational AutoEncoders* (VAEs) [1] or *Generative Adversarial Networks* (GANs), is used to solve the problem of learning a joint distribution over all the variables. As a result, these models can simulate how data is created, contrary to discriminative modeling, which learns a predictor given the observations [2]. Hence, multiple applications have been developed using generative modeling, while GANs and VAEs are usually employed for image analysis [3][4].

VAE is a deep generative model [5] that can be viewed as a combination of coupled encoder and decoder Bayesian networks that were independently parameterized. The first network (encoder) is the recognition model responsible for mapping the input data to a latent vector, delivering an estimate of its posterior over latent random variables. The result is a model capable of providing an amortized inference procedure for the computation of latent representations. In contrast, the second network (decoder) is the generative model that reverses the process. At the same time, the decoder can estimate an implicit density model given the data. Therefore, according to the Bayes rule, the recognition model is the approximate inverse of the generative model [2][6].

VAEs enhance the conventional autoencoder by introducing a Bayesian component that can learn parameters to represent the probability distribution of the data. Such follows the premise that a given training set $x$ can be generated from an underlying unobserved representation $z$ given the selected prior density function P($z$) [7]. The recognition model construction has a crucial characteristic as it can be arbitrary complex and still use a single feedforward pass from input to latent variables, providing a fast process. Conversely, several main problems are associated with VAEs [2][6]: the sampling procedure will induce noise in the gradients required, and generated data, in the case of images, tend to look blurry with a high probability of looking unrealistic; the usual approach of employing a Gaussian distribution as prior typically limits the model to learn unimodal representations, without allowing dissimilar or mixed data distributions; models are prone to suffer from the curse of dimensionality; also latent representation lacks an interpretable meaning.

This work addresses the last stated problem. Particularly, the main goal of this work is to provide an interpretable model that can then be used as feature creation for classification in a probabilistic classifier. The rationale behind the proposed solution is to provide a methodology capable of filtering uninformative features. To address this challenge, the proposed solution uses *Convolutional Neural Networks* (CNNs) for both recognition and generative models, producing an individual representation for each kernel of the last convolution layer of the recognition model. As a result, a different set of latent random variables was assessed for each of these kernels. The VAE was trained using the standard unsupervised learning methodology. Afterward, the encoder was employed as a feature creation layer (with frozen weights) for the probabilistic classifier, trained using supervised learning. The models examined data from the standard MNIST dataset (in both training procedures) as proof of concept.

Two approaches were examined for selecting which set of latent variables is more relevant for the probabilistic classifier. It is worth noting that each latent variable was sampled from a distribution associated with a kernel of the last convolution layer of the encoder. A distribution was created for each kernel; hence, each latent variable represents one of the kernels, and these variables were used as features for the classification procedure. As a result, using feature selection on the sampled latent variables leads to kernel selection of the last convolution layer of the encoder. The first feature selection approach used a wrapper method, employing *Sequential Forward Selection* (SFS) to choose the most important kernels. In contrast, the second used a filter method to determine the relevance of each kernel, ranked according to which are the most dissimilar kernels. This second proposed method follows the main hypothesis of this work that an interpretation of the latent representation can be performed by examining the shape of the created distributions, suggesting that the least relevant distributions can be identified by how similar they are to the other distributions. As a result, it is possible to filter the uninformative features (when the encoder is used for feature creation) by removing the irrelevant distributions. In this view, SFS was used as a benchmark for the proposed second method.

The proposed technique follows the rationale of model pruning, where the goal is to make smaller models that focus only on the most relevant aspects, leading to compact, memory-efficient modules. This is essential for building applications for resource-constrained devices, frequently used in the *Internet of Things* (IoT) [8] [9]. The following sections present the developed methods, followed by the results and discussion. It is then concluded by presenting the main conclusions.

## 2. Developed Methods

The developed methodology is composed of two main steps. The first refers to producing a VAE that can create an individual representation for each kernel of the last convolution layer of the recognition model. This model maps $x$ to posteriors distributions $Q(z|x)$ in the encoder while the likelihood $P(x|z)$ is parametrized by the decoder. The second step involved using the encoder model as feature creation (with frozen weights), employing a feature selection procedure to filter features from uninformative kernels. Such produced a subset $u$ that fed the probabilistic classifier to make a forecast of the labels $y$. Fig. 1 presents a simplified overview of the proposed methodology. All the code was developed in Python 3 using TensorFlow.

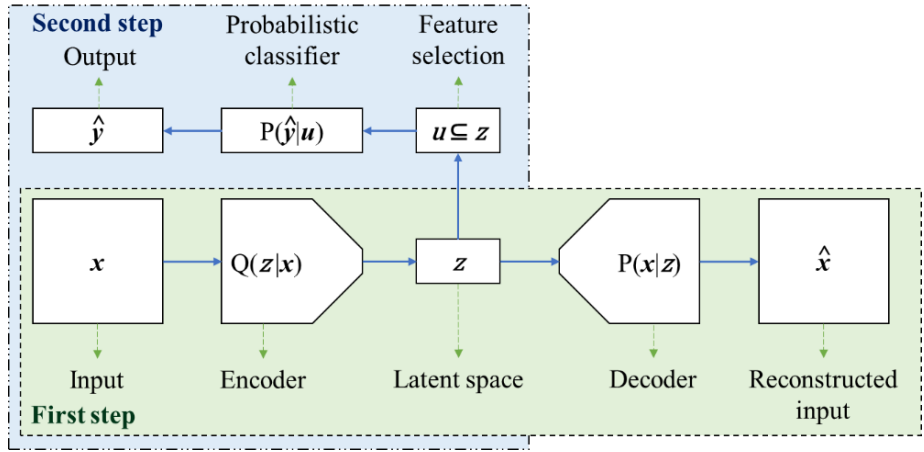

**Fig. 1.** Simplified overview of the proposed methodology, composed of two main steps. The first step develops the VAE, whose encoder was then used for feature creation that fed the classifier created in the second step.

### 2.1. Variational Autoencoders

The standard VAE is a latent variable model where the decoder assumes a Gaussian probability distribution whose parameters (mean and variance) are provided by the latent vectors. This assumed normal probability distribution could be described as [10]

$$P(x|z) = N\big(x|G_\mu(z), G_\sigma(z)^2 \times I\big), \tag{1}$$

where the first and second terms of the distribution are the mean and the variance of the latent vector, respectively. $I$ is an identity matrix of a suitable size. Through the reparameterization process, $z$ can be estimated by

$$z = \mu(x) + \sigma(x) \times \varepsilon;\ \varepsilon \sim N(0,1). \tag{2}$$

This process allows the differentiation of functions that involve *z* concerning the parameters of its distribution (in this case, mean and variance), enabling backpropagation.

During training, the optimization objective is usually related to the *Evidence Lower Bound* (ELBO), which uses the *Kullback-Leibler Divergence* (KLD) to measure closeness from the distribution Q to a target distribution. Maximizing the ELBO will maximize the log-marginal likelihood of the decoder. However, it is generally preferred to minimize the negative ELBO defined by [10] [11]

$$-\text{ELBO} = \text{KLD}[Q(\boldsymbol{z}|\boldsymbol{x})||P(\boldsymbol{z})] - E_{\boldsymbol{z}\sim Q(\boldsymbol{z}|\boldsymbol{x})}[\log[P(\boldsymbol{x}|\boldsymbol{z})]]. \tag{3}$$

This cost function comprises two parts, a regularization term and a reconstruction loss. The first is a KLD regularizer, a distance function that leads the optimization procedure to search for solutions whose densities are close to the prior on the latent representation, defined by P(*z*). The second part evaluates the probability of the data given the model and can be assessed by the cross-entropy of the VAE input and the output data. As the reparameterization trick is used, when the model is based on Gaussian distributions, the first part computation can be carried out by [6]

$$\text{KLD}[N(\mu(\boldsymbol{x}), \sigma(\boldsymbol{x}))||N(0,1)] = -\frac{1}{2}\sum[1 + \log(\sigma(\boldsymbol{x})) - \mu(\boldsymbol{x})^2 - e^{\sigma(\boldsymbol{x})}]. \tag{4}$$

It is important to notice that a standard Gaussian is a typical choice for the prior as it encourages the encoder encodings to be evenly distributed around the center of the latent space.

2.2. Implemented Variational Autoencoder Architecture

The CNN is an excellent base model for the encoder since its local connectivity characteristics allow the network to recognize local patterns in the input data. This network usually employs a combination of convolution and pooling layers to implement a transformation of the inputs while reducing the redundancy [12]. The decoder must carry out the opposite operation, which can still use a CNN but with deconvolutional layers that associate a single input activation to various outputs. Unpooling can also be used to help in the procedure. As a result, the network output is enlarged compared to the compressed input latent vector [13].

A standard architecture was used as a proof of concept in this work. In particular, a variation of LeNet-5 was employed for the encoder, composed of a convolution layer, followed by a pooling layer which, in turn, was followed by a convolution layer. Each kernel output of this last layer was then flattened and fed the sampling layer. This sampling layer estimates the two parameters of the Gaussian distribution and samples one value which is part of the array that defines the output latent vector. As a result, a Gaussian distribution is produced for each of the last convolution layer's kernels, and a value is sampled from this distribution. The decoder output shape should match the encoder input data. Hence, the latent vector fed the decoder' dense layer (a standard fully connected layer frequently used in feedforward neural networks). The number of neurons used in this dense layer was chosen in a way to avoid the need to use unpooling layers, simplifying the model architecture. The output of this dense layer was then reshaped and fed a sequence of three deconvolutional layers that generated the desired output shape (equal to the encoder input shape).

The implemented VAE architecture is presented in Fig. 2. The layer activation function was selected to be the *Rectified Linear Unit* (ReLU) to improve the network training time. Likewise, during training, the reconstruction loss was the tracked metric, and the model early stopped, with a patient of 5 epochs from the possible 100 epochs, if a minimum improvement of 1 in the tracked metric was not met.

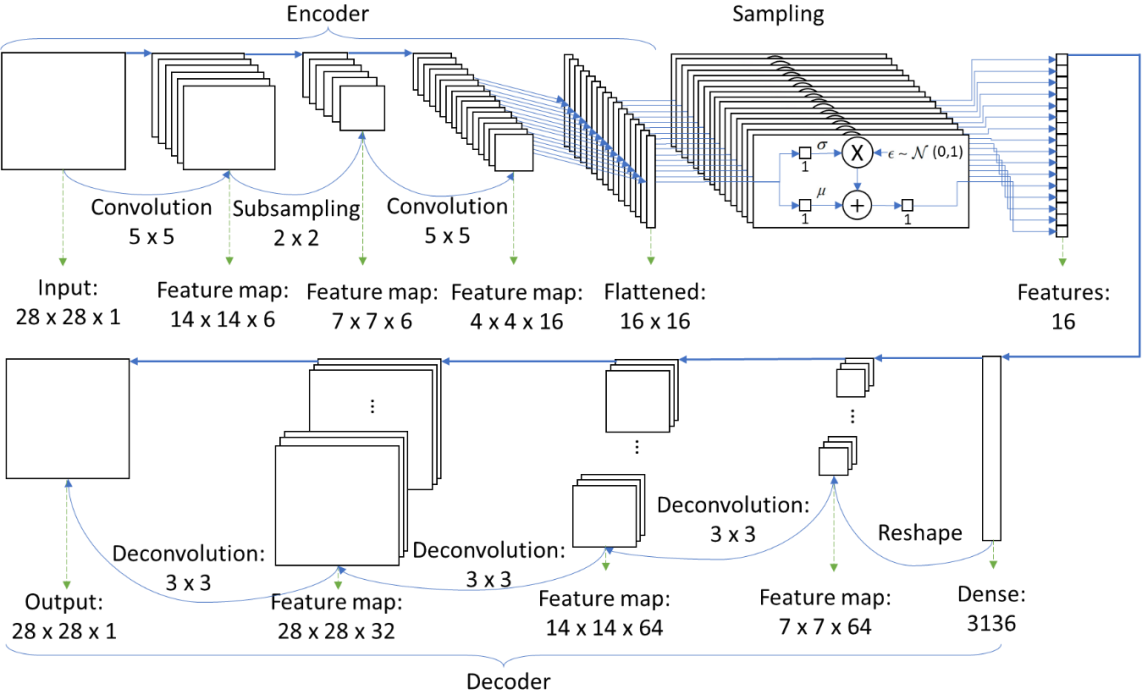

**Fig. 2.** Implemented VAE architecture for the two-dimensional input. The operations employed by the layers and their produced output shape are presented.

## 2.3. Kernel Selection Methods

The standard VAE implementation uses one dense layer to learn each parameter of the employed distribution. Hence, two dense layers are usually used since two parameters define the Gaussian distribution. The number of neurons on these layers is the same and defines the latent space. All neurons of the two dense layers are connected to all kernels of the previous convolutional layer, so they all learn distributions from the same feature data. The proposed approach of this work goes in the opposite direction, using two dense layers (with just one neuron in each) connected to each kernel from the convolutional layer that provides the features to the distributions. This way, each distribution learns from a single kernel (instead of learning from all kernels), allowing the possibility of assessing which are the most relevant kernels (latent space interpretation) for classification. Such can confirm the basic assumption of this work that the sampled features (from each individual kernel) produce dissimilar output distributions (by the sampling layer) for input data from different classes.

Two methodologies were studied to select the most relevant set of latent variables. The first was the benchmark and depended on the classifier, using the standard wrapper SFS method. Particularly, two vectors were produced, one containing all latent variables and a second that was empty. The goal was to create a feature vector to provide the maximum *Accuracy* (Acc) for the intended classification. In the first iteration, all features of the first vector were used one by one alone, and the most relevant feature (the one that, when used alone, led to the highest Acc) was identified, being moved to the second vector. In the subsequent iterations, the algorithm looked for the features in the first vector that, when combined with the features in the second vector, provide the highest Acc, being sequentially moved to the second vector. This iterative process was repeated until the first vector was empty.

The first methodology is usable when the number of latent variables (denoting the number of kernels) to be evaluated is considerably small since it requires numerous simulations. This problem is addressed by the second methodology proposed in this work, which uses a filter method to assess the relevance of the features. The second method was named KLD selection (KLDS), and its pseudocode is presented in Algorithm 1. The premise was that a kernel with a lower relevance was more similar to all other kernels; hence, it was less relevant as it conveyed less unique information. Therefore, the created features could be ranked according to the kernel relevance measure, sorting them in a vector from more to less dissimilar (i.e., first was the feature with the highest kernel relevance, then the feature with the second highest kernel relevance, and successively until the feature with the lowest kernel relevance). The second methodology's algorithm is composed of three parts. In detail, in the first part of KLDS, several Monte Carlo samples were performed. For each class, the training dataset samples fed the encoder, and the latent space features were stored. As each feature represents one of the kernels in a known sequence, it was then possible to produce the distribution of that kernel for each dataset class. Therefore, *number_of_kernels* defines the number of distributions (one for each kernel) to be assessed for each dataset class (*number_of_classes*). Afterward, for the second part, the distributions assessed for all kernels of one class were examined, calculating for each kernel distribution the KLD between their distribution and the distribution of all other kernels (in pairs). These calculations were then used to determine the mean value of all KLD estimations of each kernel. This process was then repeated for all dataset classes, leading to *number_of_kernels* average KLD estimations (one per kernel), indicated by *KLD_measures*, for each class. Finally, each kernel's assessed *KLD_measures* were averaged (resulting in *KLD_kernels*) for all classes in the third. The resulting value (for each kernel) was named kernel relevance.

**Algorithm 1.** Pseudocode for the KLDS algorithm.

**inputs:** *number_of_classes*, *number_of_kernels*, *train_data*, *train_labels*, encoder
**output:** *kernel_relevance*
**for** *class* ← 1 **to** *number_of_classes* **do**
| *samples* ← encoder(*train_data*(*train_labels* **equal to** *class*))
| *kernels_distributions* ← distribution(*samples*)
| **for** *testing_kernel* ← 1 **to** *number_of_kernels* **do**
| | **for** *comparing_kernel* ← 1 **to** *number_of_kernels* **do**
| | | *KLD_measures*(*testing_kernel*, *comparing_kernel*) ← KLD(*testing_kernel*, *comparing_kernel*)
| | *KLD_kernels*(*testing_kernel*,*class*) ← mean(*KLD_measures*(*testing_kernel*,**all**))
**for** *testing_kernel* ← 1 **to** *number_of_kernels* **do**
| *kernel_relevance*(*testing_kernel*) ← mean(*KLD_kernels*(*testing_kernel*,**all**))
*kernel_relevance*(*testing_kernel*) ← sort_descending(*kernel_relevance*)

The rationale for checking these two methodologies is to verify if a classifier-independent method (the second proposed methodology) can attain a similar performance as a classifier-dependent method (the first proposed methodology). Frequently, a classifier-dependent method can produce a better performance but require the selection to be redone when the classifier changes and takes much more time to perform the selection [14]. Hence, when the number of kernels is too high, the classifier-dependent methods become unfeasible, leading to the need for the proposed classifier-independent approach.

2.4. Classification Procedure

After training the VAE, the encoder was used to produce features, and feature selection was employed to choose the most relevant features to feed a classification procedure. This classifier, whose structure is presented in Fig. 3, was composed of three dense layers, following the LeNet-5 architecture. However, these layers were probabilistic to express the uncertainty that arises from inherent data noise and the uncertainty associated with the incompleteness of the model. These two uncertainties are named aleatoric and epistemic, respectively [15].

The probabilistic classifier was developed using *Variational Inference* (VI) to train the dense layers' weights while the encode weights were frozen. These dense layers employ the flipout estimator to decorrelate the gradients within a mini-batch, resolving a central problem associated with VI weight perturbation techniques, where there is a high variance in the gradient estimates as all mini-batch samples share the same perturbation. Therefore, the output of the dense layer forward pass is given by [16]

$$Y = \Phi[X\overline{W} + [(X \circ B)\widehat{\Delta W}] \circ A] \tag{5}$$

where ∘ is the element-wise multiplication, $\Phi$ is the employed activation function (selected to be ReLU except for the output layer, which used softmax), $X$ are the samples, $\overline{W}$ are the average weights, $\Delta W_s$ is a stochastic perturbation, and two random vectors (A and B) were sampled uniformly from the –1 to 1 range. The VI was also used to solve the ELBO, and the ADAM algorithm was employed to train the model.

2.5. Performance Metrics

The performance of the classifier was assessed using Acc as the standard dataset (MNIST) has a balanced number of samples per class. The main goal is to observe if this metric changes as the number of used features vary. This metric's 95% *Confidence Interval* (CI) was also assessed.

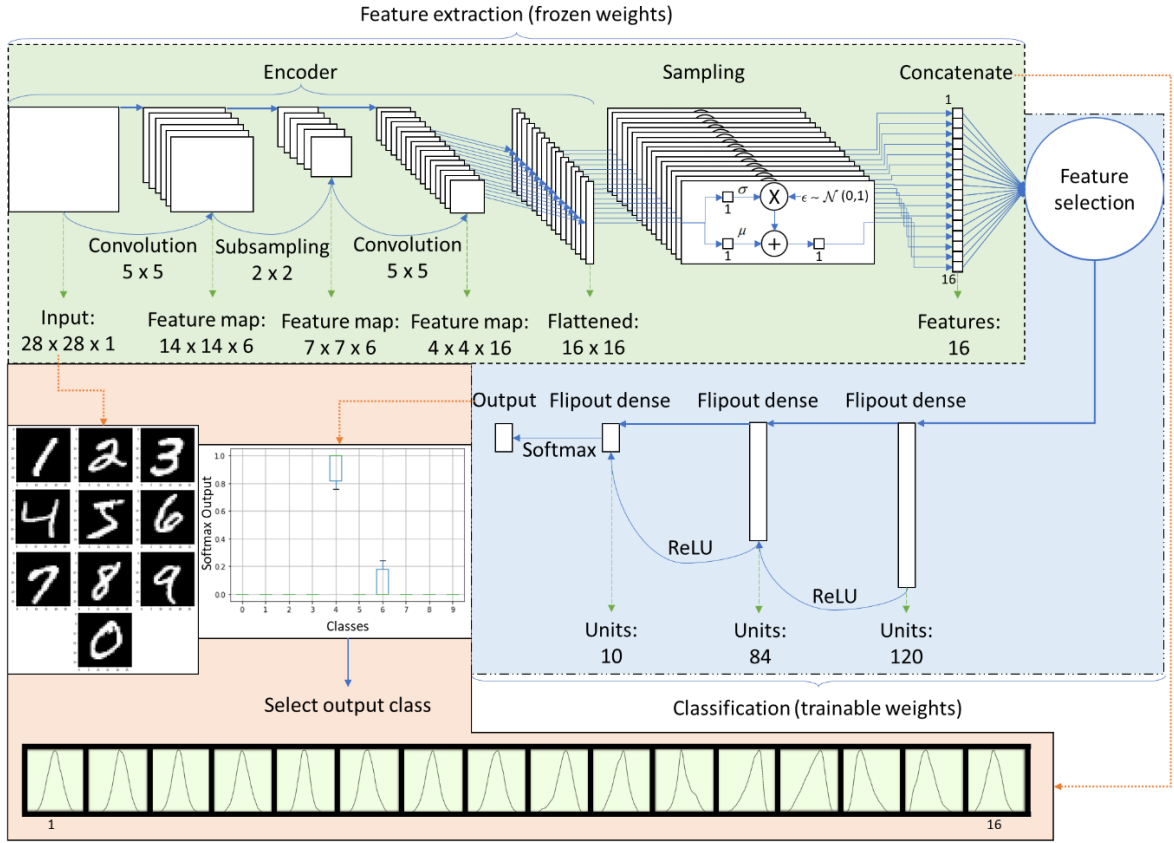

**Fig. 3.** Structure of the classifier used for the classification analysis. The feature extraction part comprises the encoder developed by the VAE, and the weights of this part were frozen (non-trainable). A feature selection procedure was employed for the classification, and the classifier's weights were optimized using supervised learning.

The quality of the reconstructed images was also examined as the number of selected features that feds the decoder varies. For this purpose, the *Feature-based SIMilarity index* (FSIM) was employed to compare the structural and feature similarity between restored and original images. This metric is based on gradient magnitude and phase congruency. The optimal similarity occurs when the FSIM is 1 [17].

## 3. Results and Discussion

This section first presents the procedure followed for validating the proposed methodologies and then further discusses the attained results.

### 3.1 Experimental Procedure

In the first experiment, the model shown in Fig. 2 was implemented, provided with data from the MNIST dataset, in which the data provider previously established training and testing datasets. Afterward, the model depicted in Fig. 3 was employed to perform the feature-based analysis with a classifier. During the training of the models, an early stopping procedure was employed to avoid overfitting, considering 20% of the training dataset as validation and a patience value of 10. The model with all kernels was further examined for the second experiment to perform uncertainty and classification analysis. The last experiment used the model presented in Fig. 2. However, the number of features fed to the decoder varied, allowing the FSIM between the input and produced images to be assessed for different settings.

3.2 Kernel Examination and Selection

The model presented in Fig. 2 was initially trained, and the goal was to evaluate the distribution of the 16 features (produced by the encoder) when only samples of one of the classes (in the test dataset) are presented. These features correspond to the distributions created by evaluating the kernels (one feature per kernel). The boxplots of the samples generated by each feature's distribution for each dataset class are presented in Fig. 4. By examining this figure, it is notorious that several distributions exhibit almost no alteration for all dataset classes. Therefore, this work hypothesizes that these kernels can be removed from the model, leading to a pruning concept. On the other hand, it was proposed that the kernels that change the most can be the most relevant for a classification task.

a)

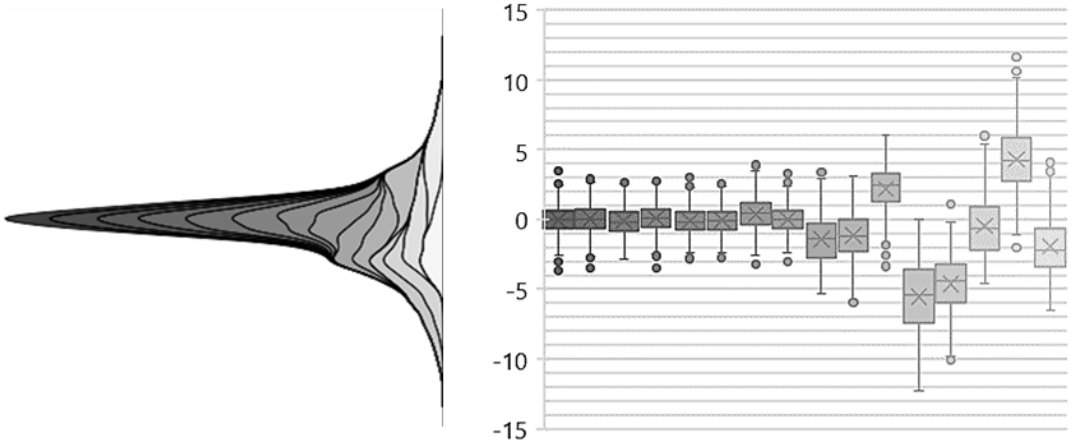

b)

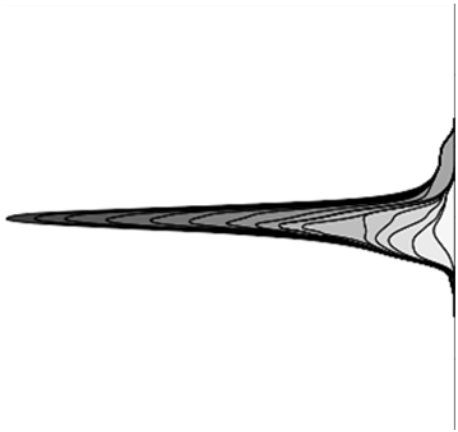 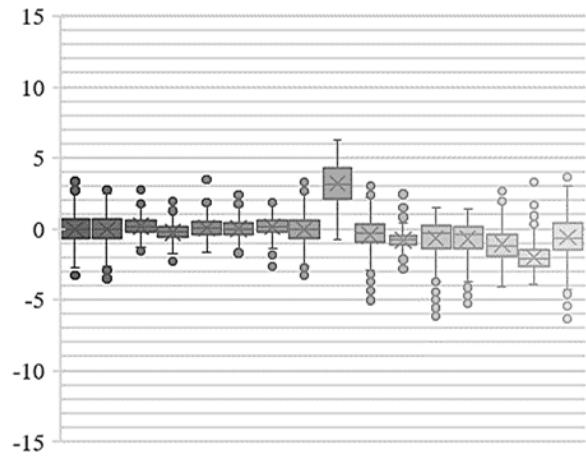

c)

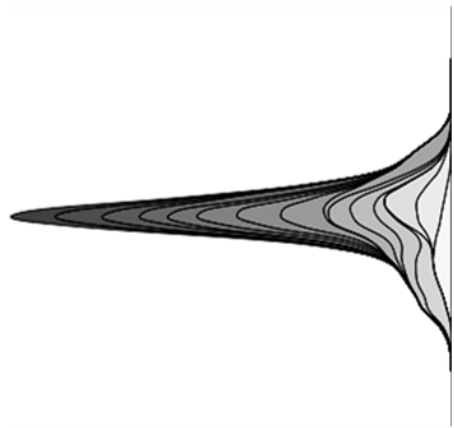 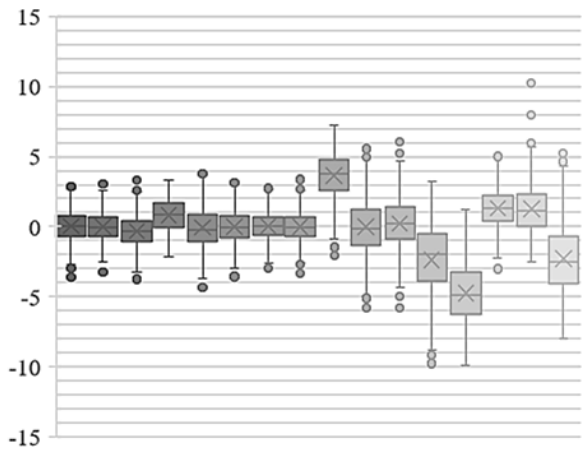

d)

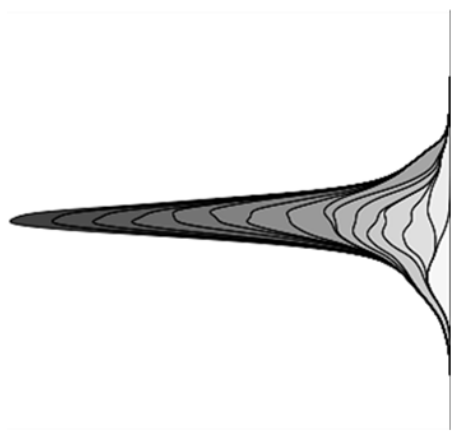 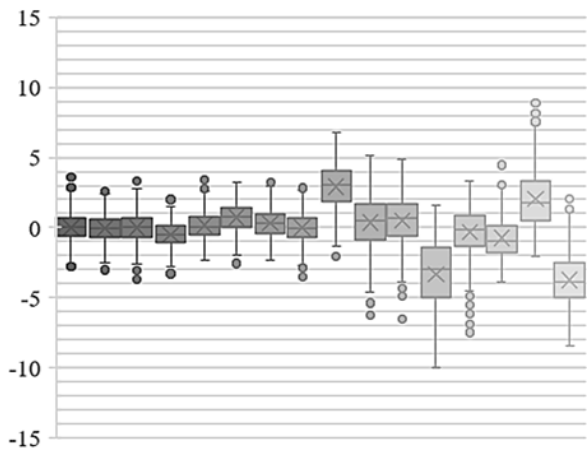

e)

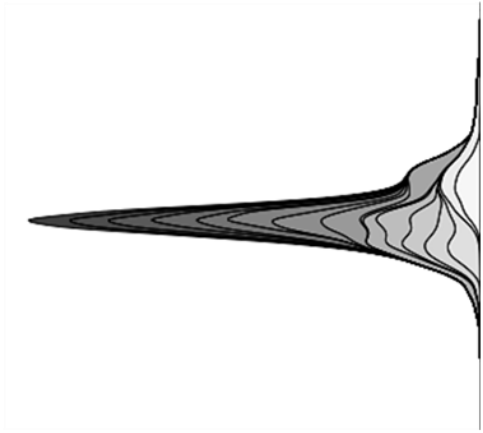 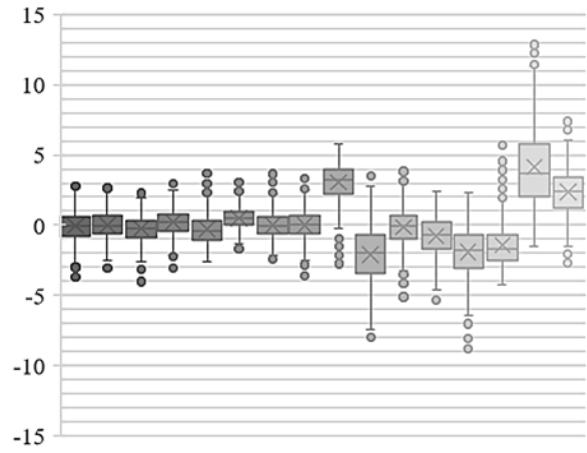

f)

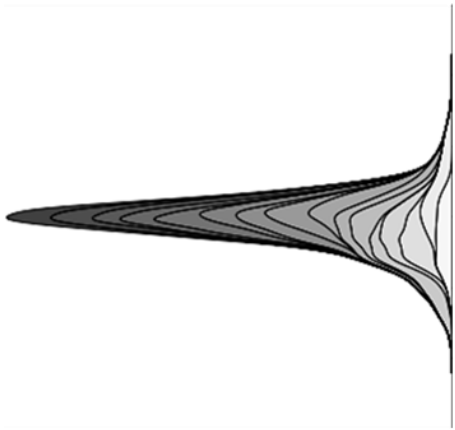 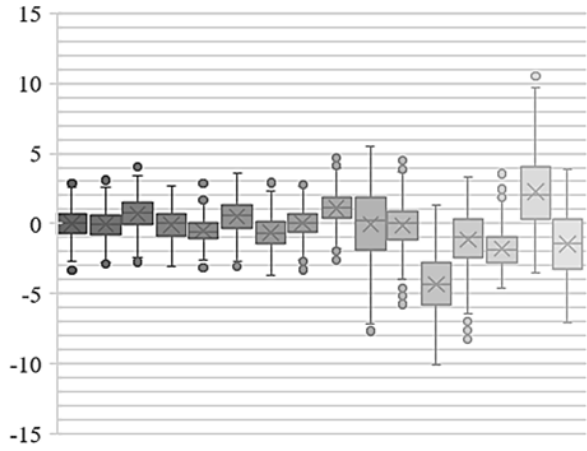

g)

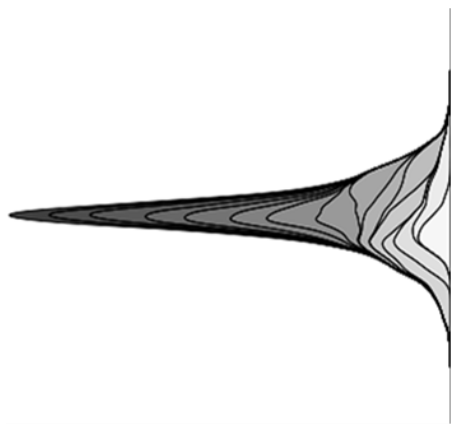 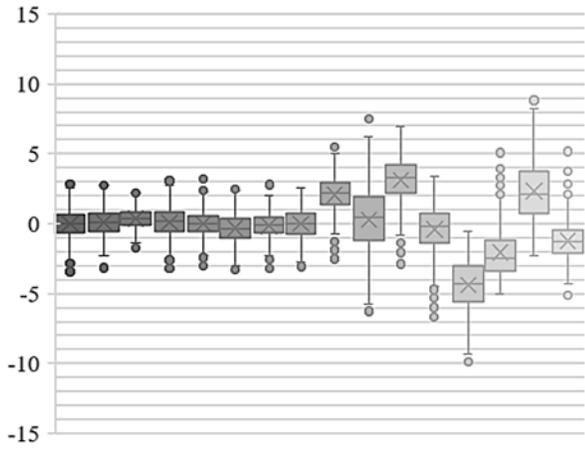

h)

i)

j)

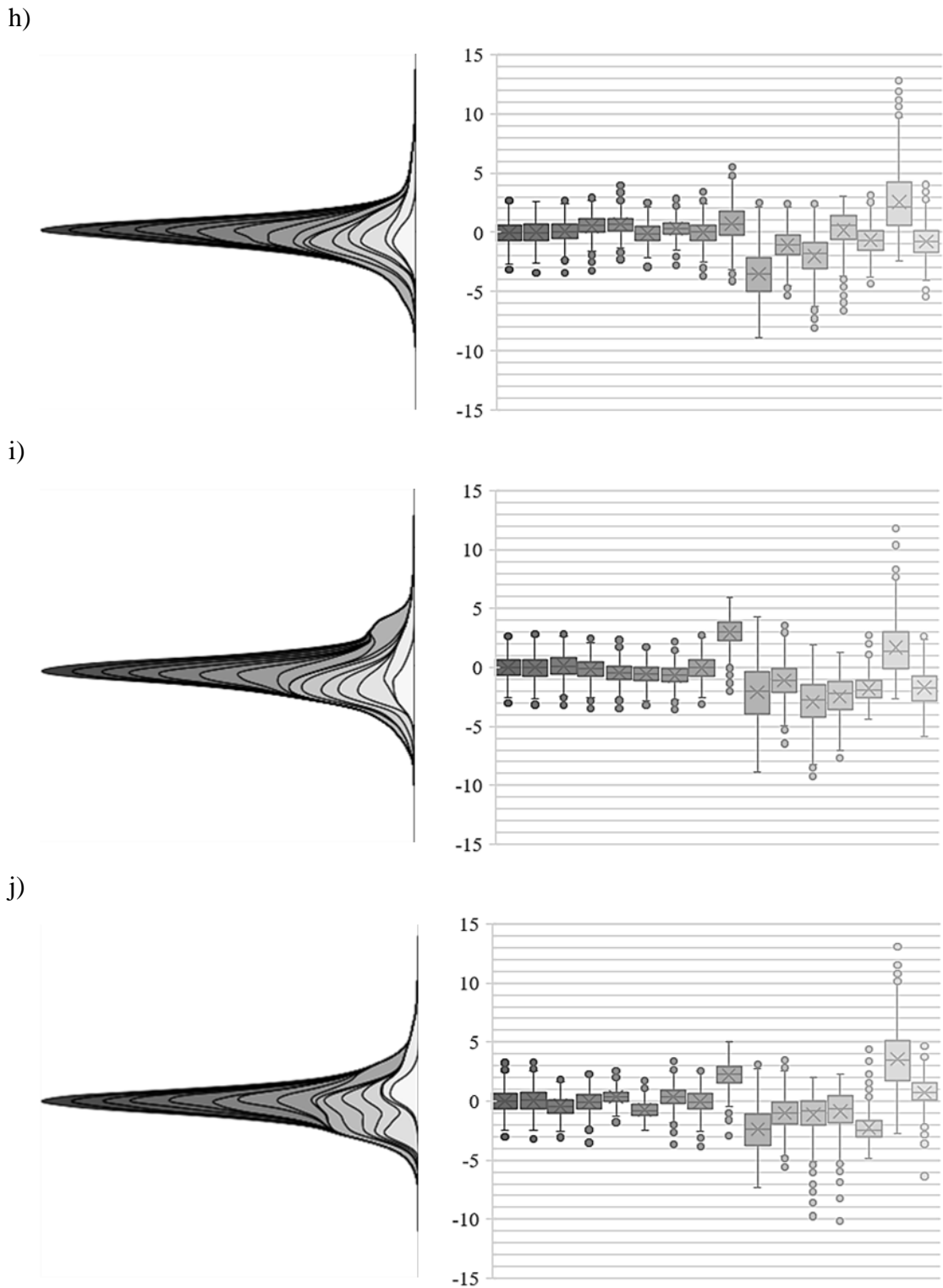

**Fig. 4.** Distributions produced by the samples of the kernel of the last convolution layer of the encoder, for the dataset classes a) 1 to j) 10, showing the 16 kernels in sequence, from left (kernel 1) to the right (kernel 16). The left figure shows the shape of each distribution

produced from the samples, whose amplitude was gradually reduced (from the first to the last kernel) to allow the visualization of all distributions. In contrast, the right figure shows the box plot of the samples.

The next step was the creation of the classifier, using the encoder (with frozen weights) as the feature creation layers and training only the probabilistic layers, using the architecture presented in Fig. 3. The Acc attained through the iterations of the evaluated feature selection algorithms is shown in Fig. 5. It was observed that both algorithms attained a similar performance, choosing the same 7 kernels as the least relevant, with almost identical order. However, the 9 most relevant differ in the selected order. Nevertheless, it is possible to observe that the last 5 selected kernels have a small contribution to the performance improvement, with only 1% variation.

As a result, the two main hypotheses of this work were confirmed, as the kernels with the lowest alteration in the kernels distributions from the dataset classes were identified as less relevant by both SFS and KLDS. The opposite was true for kernels' distributions that varied the most among the dataset classes. However, it is essential to observe that the KLDS only required 1 test per iteration, leading to 16 examinations. At the same time, SFS made 136 examinations over the 16 iterations. Hence, KLDS only made 12% of the simulations performed by SFS while achieving the same performance, supporting the relevance of this method, especially for models with a large number of kernels to be assessed.

Another relevant aspect is the total number of parameters of the model. As presented in Fig. 5, by increasing the number of kernels, the total number of parameters used by the model also increases in a linear progression. Therefore, it is notorious that keeping only the most relevant kernels in the classification can help build models with high relevance for low computational resource devices, such as the ones used for IoT. In this work, if we remove the 5 least relevant kernels, the total number of parameters of the model is reduced by 4% (1,200 parameters) compared to using all 16 kernels. This reduction can be substantial for more complex models that employ millions of parameters. As an example, using TensorFlow lite to convert the model (with deterministic layers in the classification) without changing the quantization but removing the 5 less relevant kernels can lead to a reduction of a of almost 25 KB in a microcontroller program, a value that can be relevant for resource-constrained, where the program will include multiple programming modules in addition to the machine learning model.

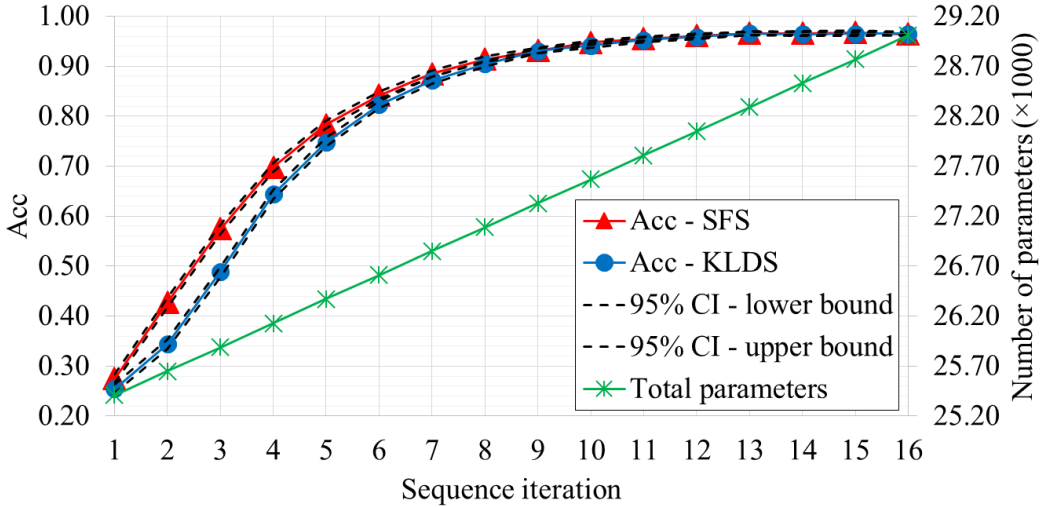

**Fig. 5.** Variation of the Acc through the sequence iteration of the evaluated feature selection algorithms. The total number of parameters of the model, as the number of kernels varies, is also presented.

3.3 Uncertainty and Classification Analysis

A relevant aspect of the employed probabilistic classifier is that it can express epistemic and aleatoric uncertainties. The first is of particular interest as it represents the uncertainty of the model. A Monte Carlo sample was used to measure this uncertainty, sampling the forecasts of the model multiple times for each example of the test dataset. An example of the variation in the models' forecast (the softmax output) from a specific example that was misclassified is presented in Fig. 6. It is notorious that choosing which class should be the output is problematic. Hence, the prediction of the model for this example should not be trusted.

Furthermore, it is relevant to notice a difference in performance when the classifier used probabilistic layers and when these were changed to deterministic layers. It was observed that the probabilistic model attained an accuracy (in the test dataset) of 97%, while the performance was 1% lower for the deterministic model. These results suggest that combining probabilistic layers with the VAE's encoder can better learn the relevant patterns than a deterministic model. Another important aspect is the performance comparison with a classifier whose encoder was developed using the standard VAE model, which uses two dense layers to learn the Gaussian distribution parameters, and all distributions are connected to all kernels. This classifier reached an average Acc that was 2% lower than the classifier based on the proposed approach, thus supporting the relevance of the developed methodology.

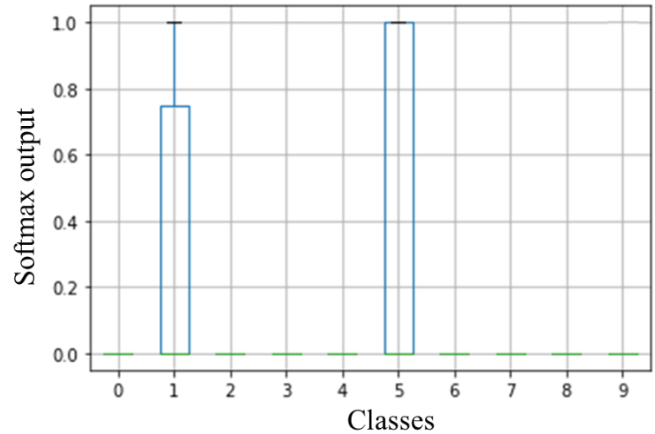

**Fig. 6.** Example of the variation in the models' forecast (the softmax output) from a specific misclassified example with high epistemic uncertainty.

3.4 Reconstruction Appraisal

To further corroborate that the kernels found to be less relevant (kernels whose distribution varies less among the dataset classes) can be removed, the sequence chosen by KLDS was examined using FSIM to assess the quality of the reconstructed images. The results of this analysis are presented in Fig. 7, which shows the FSIM for all dataset samples. A particular sample was also displayed as an example showing the progress in the reconstruction as the number of used kernels increases (adding the sequence from most to less relevant).

By examining Fig. 7, it is possible to conclude that the 7 least relevant kernels did not contribute to a relevant improvement in the FSIM, and that for some samples, an even lower number of kernels could be used (as the example presented in the figure). These results substantiate the previous analysis.

**4. Conclusions**

Kernel selection approaches for classifiers based on VAE's encoder (developed with a CNN) are presented in this work, following the rationale of model pruning. The proposed techniques were revealed to be capable of exploring the VAE latent space with an interpretation based on the produced kernels' distributions, addressing one of the identified main problems associated with VAEs.

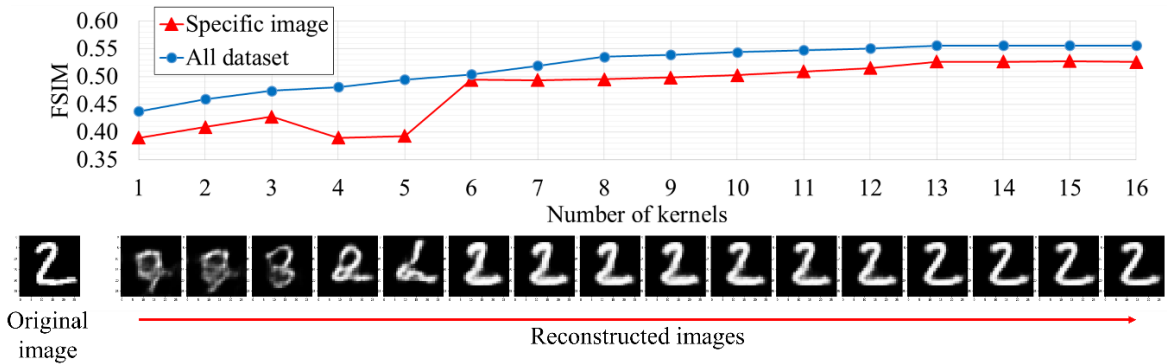

**Fig. 7.** Variation of the FSIM as the kernels selected by KLDS is progressively used (one by one), from more to less relevant. The image also displays the FSIM variation for a specific example (original image) and the progression in the reconstructed images as the number of used kernels increases.

The main hypothesis of this work was confirmed as it was observed that some kernels do not convey relevant information and can be removed. Particularly, the proposed KLDS algorithm showed that the kernels whose distributions are more dissimilar are likely the most relevant. It was also observed that when the 5 least relevant kernels were removed, both SFS and KLDS reached the same performance, validating the relevance of the later algorithm.

Regarding performance, the examined classifier with probabilistic layers was found to be superior to a model with the same architecture but using deterministic layers. This result is of utmost relevance as probabilistic models can express the epistemic uncertainty, which can be useful in areas such as clinical diagnosis [18], where it is critical to know the confidence of the classifier. As a result, the proposed methodology can be used for building applications for resource-constrained devices in multiple IoT areas, such as human activity analysis [19], where there is a need to address the bottleneck in memory usage of the devices to allow the machine learning models to be implemented on small microcontroller units [20]. Therefore, the next steps of this research should further validate the attained results in real-world data.

**Acknowledgments**

This research was funded by LARSyS (Project - UIDB/50009/2020).

The authors acknowledge the Portuguese Foundation for Science and Technology (FCT) for support through Projeto Estratégico LA 9 – UIDB/50009/2020, and ARDITI (Agência Regional para o Desenvolvimento da Investigação, Tecnologia e Inovação) under the scope of the project M1420-09-5369-FSE-000002 – Post-Doctoral Fellowship, co-financed by the Madeira 14-20 Program - European Social Fund.

**References**

[1] D. Kingma, M. Welling, Auto-encoding variational bayes, in: 2nd International Conference on Learning Representations, ICLR 2014, Banff, Alberta, Canada, 2014.

[2] D. Kingma, M. Welling, An Introduction to Variational Autoencoders, Foundations and Trends® in Machine Learning. 12 (2019) 307–392.

[3] S. Kazeminia, C. Baur, A. Kuijper, B. Ginneken, N. Navab, S. Albarqouni, A. Mukhopadhyay, GANs for medical image analysis, Artificial Intelligence in Medicine. 109 (2020) 101938.

[4] L. Ternes, M. Dane, S. Gross, M. Labrie, G. Mills, J. Gray, L. Heiser, Y. Chang, A multi-encoder variational autoencoder controls multiple transformational features in single-cell image analysis, Communications Biology. 5 (2022) 1–10.

[5] D. Rezende, S. Mohamed, D. Wierstra, Stochastic backpropagation and approximate inference in deep generative models, in: Proceedings of the 31st International Conference on Machine Learning, Bejing, China, 2014.

[6] R. Wei, C. Garcia, A. El-Sayed, V. Peterson, A. Mahmood, Variations in Variational Autoencoders - A Comparative Evaluation, IEEE Access. 8 (2020) 153651–153670.

[7] L. Chen, S. Dai, Y. Pu, E. Zhou, C. Li, Q. Su, C. Chen, L. Carin, Symmetric Variational Autoencoder and Connections to Adversarial Learning, in: Proceedings of the 21st International Conference on Artificial Intelligence and Statistics (AISTATS), Lanzarote, Spain, 2018.

[8] A. Ashiquzzaman, S. Kim, D. Lee, T. Um, J. Kim, Compacting Deep Neural Networks for Light Weight IoT & SCADA Based Applications with Node Pruning, in: 2019 International Conference on Artificial Intelligence in Information and Communication (ICAIIC), Okinawa, Japan, 2019.

[9] C. Qi, S. Shen, R. Li, Z. Zhao, Q. Liu, J. Liang, H. Zhang, An efficient pruning scheme of deep neural networks for Internet of Things applications, EURASIP Journal on Advances in Signal Processing. 31 (2021) 1–21.

[10] S. Lee, S. Park, B. Choi, Application of domain-adaptive convolutional variational autoencoder for stress-state prediction, Knowledge-Based Systems. 248 (2022) 108827.

[11]  D. Blei, A. Kucukelbir, J. McAuliffe, Variational Inference: A Review for Statisticians, Journal of the American Statistical Association. 112 (2016) 859–877.

[12]  J. Nagi, F. Ducatelle, G. Caro, D. Cireşan, U. Meier, A. Giusti, F. Nagi, J. Schmidhuber, L. Gambardella, Max-pooling convolutional neural networks for vision-based hand gesture recognition, in: 2011 IEEE International Conference on Signal and Image Processing Applications (ICSIPA), Kuala Lumpur, Malaysia, 2011.

[13]  H. Noh, S. Hong, B. Han, Learning deconvolution network for semantic segmentation, in: 2015 IEEE International Conference on Computer Vision (ICCV), Santiago, Chile, 2015.

[14]  S. Mostafa, F. Morgado-Dias, A. Ravelo-García, Comparison of SFS and mRMR for oximetry feature selection in obstructive sleep apnea detection, Neural Computing and Applications. 32 (2018) 1–21. https://doi.org/10.1007/s00521-018-3455-8.

[15]  S. Ryu, Y. Kwon, W. Kim, A Bayesian graph convolutional network for reliable prediction of molecular properties with uncertainty quantification, Chemical Science. 36 (2019) 8438–8446.

[16]  Y. Wen, P. Vicol, J. Ba, D. Tran, R. Grosse, Flipout: Efficient Pseudo-Independent Weight Perturbations on Mini-Batches, in: Sixth International Conference on Learning Representations, Vancouver, Canada, 2018.

[17]  L. Zhang, L. Zhang, X. Mou, D. Zhang, FSIM: A Feature Similarity Index for Image Quality Assessment, IEEE Transactions on Image Processing. 20 (2011) 2378–2386.

[18]  O. Ozdemir, R. Russell, A. Berlin, A 3D Probabilistic Deep Learning System for Detection and Diagnosis of Lung Cancer Using Low-Dose CT Scans, IEEE Transactions on Medical Imaging. 39 (2019) 1419–1429.

[19]  H. Amroun, M. Temkit, M. Ammi, Best Feature for CNN Classification of Human Activity Using IOT Network, in: IEEE/ACM Int'l Conference on & Int'l Conference on Cyber, Physical and Social Computing (CPSCom) Green Computing and Communications (GreenCom), Exeter, United Kingdom, n.d.

[20]  J. Know, D. Park, Hardware/Software Co-Design for TinyML Voice-Recognition Application on Resource Frugal Edge Devices, Applied Sciences. 11 (2021) 11073.